\documentclass[runningheads,a4paper]{llncs}
\usepackage{times}
\usepackage{latexsym}
\usepackage{amsmath}
\usepackage{multirow}
\usepackage{url}
\usepackage{float}
\usepackage{amssymb}
\usepackage{rotating}
\usepackage{comment}
\usepackage{caption}
\usepackage{subcaption}
\usepackage{color}
\def\todo#1{\bgroup \textcolor{red}{(#1)}\egroup}
\def\note#1{\bgroup \textcolor{blue}{(#1)}\egroup}
\usepackage{array}
\newcolumntype{C}{$>${\centering\arraybackslash}p{7cm}}

\def\todo#1{\bgroup \textcolor{red}{(#1)}\egroup}

\title{New word analogy corpus for exploring embeddings of Czech words}
\titlerunning{New word analogy corpus for exploring embeddings of Czech words}
\toctitle{New word analogy corpus for exploring embeddings of Czech words}

\author{Luk\'{a}\v{s} Svoboda\inst{1}\inst{2}
 \and Tom\'{a}\v{s} Brychc\'{i}n\inst{1}\inst{2}
}

\authorrunning{Luk\'{a}\v{s} Svoboda\inst{1}\inst{2}
	\and Tom\'{a}\v{s} Brychc\'{i}n\inst{1}\inst{2}
}
\tocauthor{Luk\'{a}\v{s} Svoboda, Tom\'{a}\v{s} Brychc\'{i}n}

\institute{Department of Computer Science and Engineering, Faculty of Applied Sciences, \\
University of West Bohemia, Univerzitn\'{i}­ 8, 306 14  Plze\v{n}, Czech Republic
\and NTIS---New Technologies for the Information Society, Faculty of Applied Sciences, \\
University of West Bohemia, Univerzitn\'{i}­ 8, 306 14  Plze\v{n}, Czech Republic \\
\email{\{svobikl,brychcin\}@kiv.zcu.cz} \\
\url{nlp.kiv.zcu.cz}
}

\begin{document}
\maketitle

\begin{abstract}

The word embedding methods have been proven to be very useful in many tasks of NLP (Natural Language Processing). Much has been investigated about word embeddings of English words and phrases, but only little attention has been dedicated to other languages.

Our goal in this paper is to explore the behavior of state-of-the-art word embedding methods on Czech, the language that is characterized by very rich morphology.
We introduce new corpus for word analogy task that inspects syntactic, morphosyntactic and semantic properties of Czech words and phrases. We experiment with Word2Vec and GloVe algorithms and discuss the results on this corpus. The corpus is available for the research community.

\end{abstract}

\section{Introduction}\label{sec:motivation}

Word embedding is the name for techniques in NLP (Natural Language Processing) where meaning of words or phrases is represented by vectors of real numbers.

It was shown that the word vectors can be used for significant improving and simplifying of many NLP applications \cite{Collobert08aunified,DBLP:journals/corr/abs-1103-0398}. There are also NLP applications, where word embeddings does not help much \cite{Andreas:2014}. 

There has been introduced several methods based on the feed-forward NNLP (Neural Network Language Model) in recent studies. One of the Neural Network based models for word vector representation which outperforms previous methods on word similarity tasks was introduced in \cite{Huang-2012}. The word representations computed using NNLP are interesting, because trained vectors encode many linguistic properties and those properties can be expressed as linear combinations of such vectors.

Nowadays, word embedding methods Word2Vec \cite{mikolov2013efficient} and GloVe \cite{pennington2014glove} significantly outperform other methods for word embeddings. Word representations made by these methods have been successfully adapted on variety of core NLP task such as Named Entity Recognition \cite{siencnik2015adapting,demir2014improving}, Part-of-speech Tagging \cite{al2013polyglot}, Sentiment Analysis \cite{pontiki2015semeval}, and others.

There are also neural translation-based models for word embeddings \cite{cho2014learning,bahdanau2014neural} that generates an appropriate sentence in target language given sentence  in source language, while they learn distinct sets of embeddings for the vocabularies in both languages. Comparison between monolingual and translation-based models can be found in \cite{HillCJDB14}.

Many researches have investigated the behavior of these methods on English, but only little attention has been dedicated to other languages. In this work we focus on Czech that is a representative of Slavic languages. These languages are highly inflected and have a relatively free word order. Czech has seven cases and three genders. The word order is very variable from the syntactic point of view: words in a sentence can usually be ordered in several ways, each carrying a slightly different meaning. All these properties complicate the learning of word embeddings.

In this article we are exploring whether are word embedding methods as good on highly inflected languages like Czech as they are on English. It has been shown that such word embedding models improve Named Entity Recognition on Czech \cite{demir2014improving}, but we would like to investigate if they capture the semantic and syntactic relationships independently from specific task.

There is a variety of datasets for measuring semantic relatedness between English words, such as \emph{WordSimilarity-353} \cite{ws353}, \emph{Rubenstein and Goodenough (RG)} \cite{RubensteinGoodenough65}, \emph{Rare-words} \cite{Luong-etal:conll13:morpho}, \emph{Word pair similarity in context} \cite{Huang:2012:IWR:2390524.2390645}, and many others. To the best of our knowledge, there is only one such corpus for Czech \cite{KrcmarKJ11}, which is essentially only translation of RG corpus into Czech.

Except the similarity between words, we would like to explore other semantic and syntactic properties hidden in word embeddings. A new evaluation scheme based on word analogies were presented in \cite{mikolov2013efficient}. By examining various dimensions of differences we can achieve interesting results, for example: $\textrm{vector("king")} - \textrm{vector("man")}$ is close to $\textrm{vector("queen")} - \textrm{vector("woman")}$. Based on this approach and our need to further use and investigate the word embedding methods on Czech, we have decided to build semantic-syntactic word analogy dataset. Especially, we focus on exploring how state-of-the-art word embedding methods carry semantics and syntax of words.

\section{Word embeddings methods}
The backbone principle of word embedding methods is the formulation of \emph{Distributional Hypothesis} in \cite{Firth:1957} that says \emph{``a word is characterized by the company it keeps"}. The direct implication of this hypothesis is that the word meaning is related to the context where it usually occurs and thus it is possible to compare the meanings of two words by statistical comparisons of their contexts. This implication was confirmed by empirical tests carried out on human groups in \cite{RubensteinGoodenough65,Charles2000}. 

The distributional semantics models typically represent the word meaning as a vector, where the vector reflects the contextual information of a word across the training corpus. Each word $w \in W$ (where $W$ denotes the word vocabulary) is associated with a vector of real numbers $\boldsymbol w \in \mathbb{R}^k$.  Represented geometrically, the word meaning is a point in a high-dimensional space. The words that are closely related in meaning tend to be closer in the space.

In this work we will focus on three monolingual models that produce high quality word embeddings. In general, given a single word in the corpus, these models predict which other words should serve as a substitution for this word. 

\subsection{CBOW}
\begin{figure}[ht!]
\centering

\begin{subfigure}{4cm}
           \includegraphics[width=4cm,height=4cm]{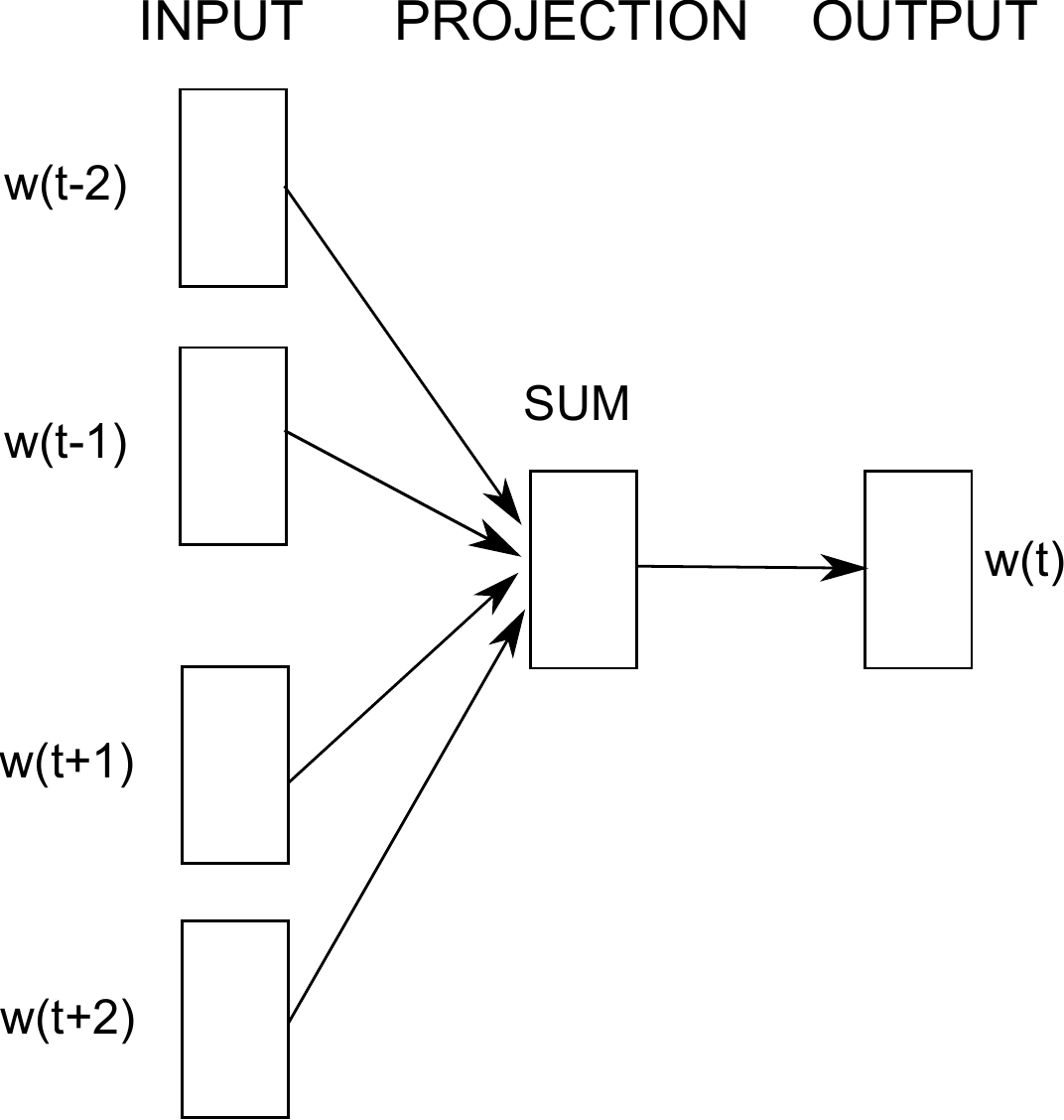}
	\caption{CBOW}
	\label{fig:cbow}
\end{subfigure}
\hspace{3mm}
\begin{subfigure}{4cm}
            \includegraphics[width=4cm,height=4cm]{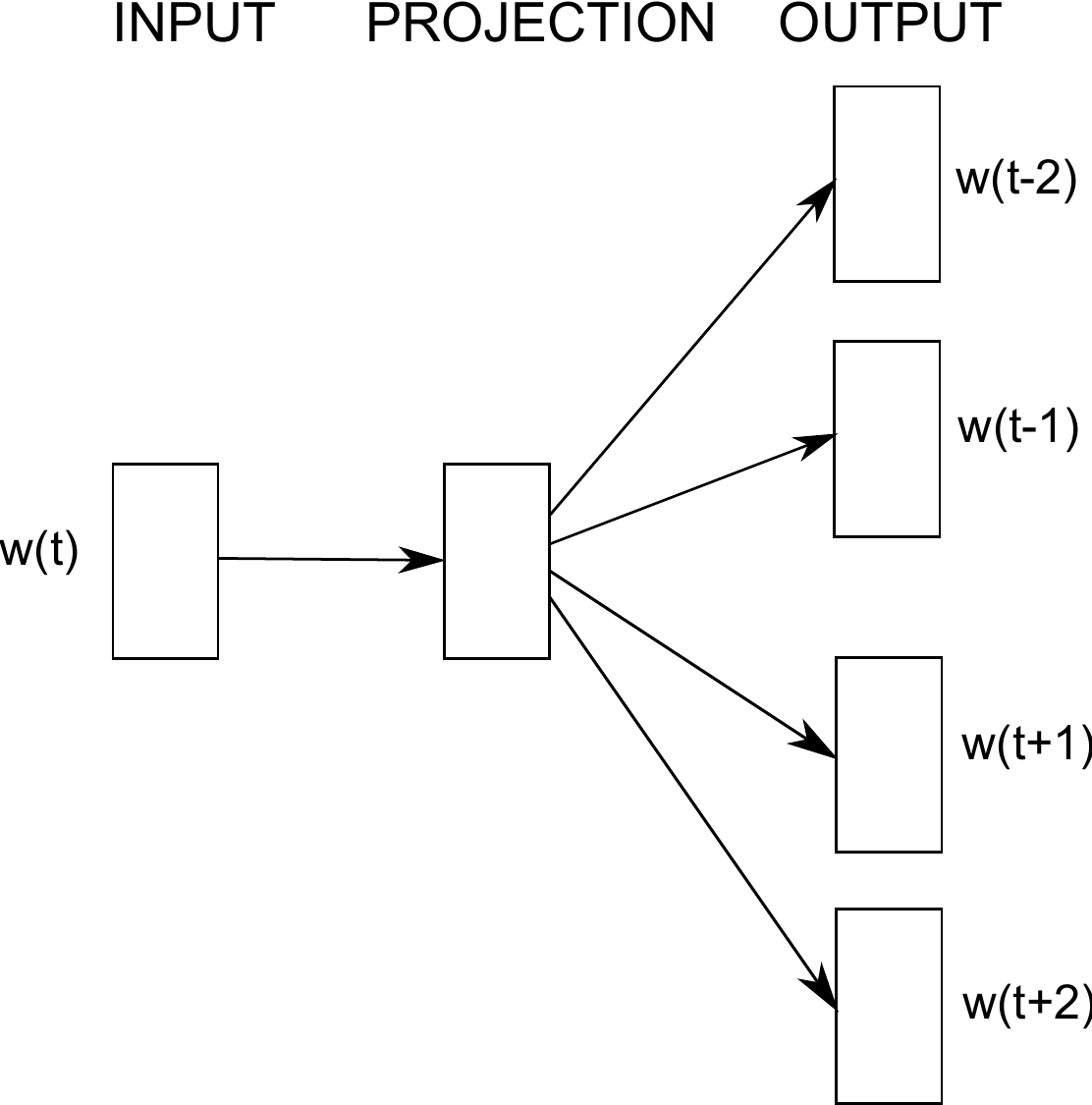}
	\caption{Skip-gram}
	\label{fig:skip}
\end{subfigure}
\caption{Nerual network models architectures.}
\end{figure}

CBOW (Continuous Bag-of-Words) \cite{mikolov2013efficient} tries to predict the current word according to the small context window around the word. The architecture is similar to the feed-forward NNLP (Neural Network Language Model) which has been proposed in \cite{bengio2006neural}. The NNLM is computationally expensive between the projection and the hidden layer. Thus, CBOW proposed architecture, where the (non-linear) hidden layer is removed and projection layer is shared between all words. The word order in the context does not influence the projection (see Figure \ref{fig:cbow}). This architecture also proved low computational complexity.

\subsection{Skip-gram}
Skip-gram architecture is similar to CBOW. Although instead of predicting the current word based on the context, it tries to predict a words context based on the word itself \cite{mikolov2013distributed}. Thus, intention of the Skip-gram model is to find word patterns that are useful for predicting the surrounding words within a certain range in a sentence  (see Figure \ref{fig:skip}). Skip-gram model estimates the syntactic properties of words slightly worse than the CBOW model, but it is much better for modeling the word semantics on English test set \cite{mikolov2013efficient} \cite{mikolov2013distributed}. Training of the Skipgram model does not involve dense matrix multiplications \ref{fig:skip} and that makes training also extremely efficient \cite{mikolov2013distributed}.

\subsection{GloVe}
GloVe (Global Vectors) \cite{pennington2014glove} model focuses more on the global statistics of the trained data. This approach analyses log-bilinear regression models that effectively capture global statistics and also captures word analogies. Authors propose a weighted least squares regression model that trains on global word-word co-occurrence counts. The main concept of this model is the observation that ratios of word-word co-occurrence probabilities have the potential for encoding meaning of words.

\section{Word analogy corpus}
In this section we present a new word analogy Czech corpus for testing word embeddings. Inspiration was taken from English corpus revealed in \cite{mikolov2013efficient}. We follow observation that the state-of-the-art models for word embeddings can capture different types of similarities between words. 
Given two pairs of words with the same relationship as a question: Which word is related to \emph{export} in the same sense as \emph{minimum} is related to \emph{maximum}? Correct answer should be \emph{import}.

Such a question can be answered with a simple algebraic operation with the vector representation of words:

\begin{equation} \label{eq:vec}
 \boldsymbol x = \textrm{vector("maximum")} - \textrm{vector("minimum")} + \textrm{vector("export")} 
\end{equation}

Difference between $\textrm{vector("maximum")}$ and $\textrm{vector("minimum")}$ should be similar to difference between $\textrm{vector("export")}$ and $\textrm{vector("import")}$. For resulting vector $\boldsymbol x$ we search in the vector space for the most similar word. When the model works well and is properly trained, we will find that the closest vector representing correct answer for our question is the vector for the word \emph{import}. 

If the model has sufficient data, it is able to learn also more complicated semantic relationships between words, such as the main city \emph{Prague} to the state \emph{Czech Republic} is with the similar relation as \emph{Paris} is to \emph{France}, or capturing the presidents of individual states, already mentioned antonyms, plural versus singular words, gradation of adjectives, and other words relationships.

To measure quality of word vectors, we have designed test set containing 8,705 semantic questions and 13,552 syntactic questions. Together, we have 22,257 combinations of questions. Dataset contains only enough frequent words on Czech Wikipedia. We split the dataset into several categories. Each category usually contains about 35--40 pairs of words with same relationship. Question has been built by all combination of word pairs in the same category.

There is a majority of word-to-word relationships, but \emph{Presidents and states} category contains also bigram-to-word (word-to-bigram) relationships such as \emph{Prague} vs. \emph{Czech Republic}.

\noindent
Semantic questions are represented in categories: 
\begin{itemize}
\item \textbf{Presidents-states-cities:} Consists of 34 pairs of states in Europe and their main cities combining 1,122 questions. There is also 1,122 questions for state with corresponding current president. 
\item \textbf{Antonyms:} This category compounds of three subcategories. In first subcategory we have 38 noun antonym pairs that is resulting in 1406 questions combined. Example of such question is: 
\emph{anode, cathode} versus \emph{export, import}. 
Similarly we have 42 adjectives pairs (such as \emph{big, small}) and 34 verb pairs - \emph{buy, sell} versus \emph{give, take}. 
\item \textbf{Family-relations (man-woman):} In this category we have 19 pairs of family representatives with man-woman relation as \emph{brother, sister} versus \emph{husband, wife}.
\end{itemize}

\noindent
Syntactic questions are represented in categories: 
\begin{itemize}
\item \textbf{Adjectives-gradation:} In this category we have two antonym pairs with three degrees of adjectives in positive, comparative, and superlative form: \emph{big, bigger} vs \emph{small, smaller}. 
\item \textbf{Nationalities (woman/man):} This category is specific for Czech language, which distinguish between masculine and feminine word relations. Every nationality has its corresponding masculine and feminine word form. For example, English word \emph{Japan} has in Czech masculine form \emph{Japonec} and feminine form \emph{Japonka}. We have 35 such pairs.
\item \textbf{Nouns-plural:} We find here 37 pairs of nouns and their plural forms.
\item \textbf{Jobs:} Category with 35 pairs of professions with masculine-feminine word relations. 
\item \textbf{Verb-past:} This category consists of verbs in present form versus verbs in past tense form, such as \emph{play, played} versus \emph{see, saw}.
\item \textbf{Pronouns:} Last category consists of pairs of pronouns in singular versus plural form. 
\end{itemize}

\section{Experiments}

In our experiments, we used unsupervised learning of word-level embeddings using Word2Vec \cite{mikolov2013efficient} and GloVe tool \cite{pennington2014glove}. 
We used the January 2015 snapshot of the Czech Wikipedia as a source of unlabeled data. 
The Wikipedia corpus has been preprocessed with the following steps: 

\begin{enumerate}
\item Removed special characters such as \emph{\#\$\&\%}, HTML tags and others. 
\item Filtering XML dumps, removed tables, links converted to normal text. We lowercase all words. We have also removed sentences with less than 5 tokens.
\end{enumerate}

The resulting training corpus contains about 2,6 billion words. For our purpose, it is useful to have vector representation of word phrases, i.e. for bigram representing state \emph{Czech Republic}, it is desirable to have one vector representing those two words. This was achieved by preprocessing the training data set to form the phrases using the \emph{Word2Phrase} tool \cite{mikolov2013distributed}.

We evaluate the word embedding models on our corpus by accuracy that is defined as
\begin{equation}
  \textrm{Acc\%} = \frac{\textrm{NC}}{\textrm{NT}}, 
\end{equation}

\noindent
where ${\textrm{NC}}$ is the number of correctly answered questions for a category and ${\textrm{NT}}$ is total number of questions in category.

In our experiments, we use \textit{cosine similarity} as a measure of similarity between two word vectors. Cosine similarity is probably the most used similarity metric for words embedding methods. It characterizes the similarity between two vectors as the cosine of the angle between them

\begin{equation}\label{eq:cos}
S_{\cos } (\boldsymbol a,\boldsymbol b)  = \frac{{\boldsymbol a \cdot \boldsymbol b}}{\| {\boldsymbol a} \| \cdot \| {\boldsymbol b} \|} = \frac{{\sum {a_i b_i } }}{{\sqrt {\sum {a_i^2 } \sum {b_i^2 } } }},
\end{equation}

\noindent
where $\boldsymbol a$ and $\boldsymbol b$ are two vectors we try to compare. The cosine similarity is used in all cases where we want to find the most similar word (or top $n$ most similar words) for a given type of analogy.

\subsection{Models settings}

During the training of Word2Vec (resp. GloVe) models, we limited the size of the vocabulary to 400,000 most frequent single token words and about 800,000 most frequent bigrams. OOV (Out-of-vocabulary) word rate was 6\%. It means that out of 22,257 questions was about 1,300 questions not seen in vocabulary. 

To train word embedding methods we use context window of size 10. We also explore results with different vector dimension (set to 100, 300, and 500). We choose to compare three training epochs as in \cite{mikolov2013efficient} for similarly sized training corpus versus ten training epochs for Word2Vec tool. For GloVe tool we choose 10 and 25 iterations, because algorithms cannot be simply compared with the same settings \cite{pennington2014glove}. Other Word2Vec and GloVe settings were on its default values.

\subsection{Results on Czech word analogy corpus}

In this section we present the accuracies for all tested models (CBOW, Skip-gram, and GloVe) on our word analogy corpus. In all tables below we present results for different vector dimension ranging between 50 and 500, except for Skip-gram model with dimension 500 and 10 training epochs, where the time of computation was much higher than with other methods. Model did not finnish after 4 days of training and results of 500 dimension vector does not substantiate such long training time.  We use notation $n$\_$D$ in the tables, $n$ means that the correct word must be between $n$ most similar words for a given analogy. $D$ denote the dimension of vectors. Accuracies are expressed in percents.

In Table \ref{tab:cbow} we present the results for CBOW model. There is a significant improvement between 3 and 10 training epochs. Interesting is also fact that 300-dimensional vectors perform better than 500-dimensional vectors on most categories. Similarly, the results for Skip-gram model are in Table \ref{tab:skip}. This model performs significantly worse on most categories in comparison with CBOW model. There is also significant overall improvement between 50-dimensional and 100-dimensional vector, but less significant between 100 and 300. Table \ref{tab:glove} shows result for GloVe model. This model gives on Czech the worst results compared to both Word2Vec models. 

Categories, where the models gives best results are \emph{Verb-past}, \emph{Noun-plural}, and \emph{State-city}. In general, all models gives better results on tasks exploring syntactic information. Poor accuracy was in categories \emph{State-presidents} and category \emph{Nationality}.

\begin{table}[ht!]
\caption{Results for CBOW.} \label{tab:cbow}
\centering
\resizebox{\textwidth}{!}{
\begin{tabular}{ l||r|r|r|r|r|r|r|r|r|r|r|r }
\hline
\multirow{2}{*}{\bf Type} &\multicolumn{12}{c}{\rule{0pt}{12pt}\bf 3 training epochs}\\
& \bf1\_50 & \bf1\_100 & \bf1\_300 & \bf1\_500 & \bf5\_50 & \bf5\_100 & \bf5\_300 & \bf5\_500 & \bf10\_50 & \bf10\_100 & \bf10\_300 & \bf10\_500\\
\hline
\bf Anton. (nouns) 	&1.35	&4.84 & 5.55 & 5.69 & 3.98 & 10.88 & 13.16& 10.95 & 5.69 & 13.44 & 16.00 & 13.30  \\
\bf Anton. (adj.) 	& 4.82 	&8.86 & 11.79 & 13.24 & 10.63 & 14.29 & 18.70& 19.16 & 13.24 & 17.31 & 23.64 & 22.76 \\
\bf Anton. (verbs) 	& 0.20 	&1.88 & 2.68& 1.25 & 2.77 & 3.13 & 6.25 & 3.57 & 2.94 & 3.84 & 7.77 & 4.38 \\
\bf State-president	& 0.00 	&0.00 & 0.18 & 0.09 & 0.18 & 0.00 & 0.98 & 0.18 & 0.45 & 0.27 & 1.43 & 0.71 \\
\bf State-city 		& 14.62 &14.8 & 16.22 & 8.47 & 29.77 & 30.93 & 32.89 & 23.26 & 35.92 & 39.57 & 42.96 & 31.82 \\
\bf Family 			& 6.42 	&9.01 & 11.60 & 9.26 & 12.10 & 17.28 & 21.85& 18.64 & 14.44 & 21.11 & 25.80 & 23.95  \\
\hline
\bf Noun-plural 	&34.46 	&42.42 & 41.74 & 44.60 & 45.95 & 53.60 & 54.35 & 54.35 & 50.45 & 57.43 & 57.43 & 57.81 \\
\bf Jobs 			&2.95	&3.87 & 3.37& 2.78 & 6.57 & 10.52 & 10.00 & 8.92 & 9.18 & 14.05 & 13.80 & 12.37 \\
\bf Verb-past 		&14.83 	&24.29 & 42.52 & 34.91 & 29.94 & 40.91 & 60.61 & 52.00 & 36.66 & 48.31 & 66.50 & 58.80\\
\bf Pronouns 		&1.59 	&3.84 & 5.95 & 3.57 & 3.97 & 8.07 & 12.70 & 10.05 & 5.69 &9.66 & 16.00 & 13.10 \\
\bf Adj.-gradation 	&12.50 	&20.00 & 22.50 & 15.00 & 20.00 & 22.50 & 22.50 & 27.50 & 20.00 & 27.50 & 25.00 & 27.50 \\
\bf Nationality 		&0.08	&0.42 & 0.33 & 0.16 & 0.84 & 0.92 & 0.84 & 1.10 & 1.26 & 1.26 & 1.26 & 2.01 \\
\hline
\multirow{2}{*}{} &\multicolumn{12}{c}{\rule{0pt}{12pt}\bf 10 training epochs}\\
& \bf1\_50 & \bf1\_100 & \bf1\_300 & \bf1\_500 & \bf5\_50 & \bf5\_100 & \bf5\_300 & \bf5\_500 & \bf10\_50 & \bf10\_100 & \bf10\_300 & \bf10\_500\\
\hline
\bf Anton. (nouns) &3.84 &	7.82 & 8.53 & 7.40 & 8.39 & 15.93 & 18.49& 16.07 & 10.38 & 19.42 & 22.76& 20.55 \\
\bf Anton. (adj.) &7.26& 11.90 & 15.45 & 15.04& 13.53 & 19.63 & 25.49& 23.58 & 16.49 & 23.05 & 30.26 & 28.92  \\
\bf Anton. (verbs) &0.89 & 1.88 & 2.86 & 3.12 & 4.01 & 5.98 & 6.43 & 6.07 & 5.09 & 6.70 & 7.59 &  7.41\\
\bf State-president &0.18 &	0.35 & 0.09 & 0.09 & 0.71 & 0.98 & 0.62 & 0.71 & 1.16 & 1.60 & 1.33 & 1.16 \\
\bf State-city &16.58 &	27.99 & 25.94 & 18.63 & 37.07 & 50.62 & 52.05 & 39.13 & 43.49 & 58.47 & 61.41 & 50.71 \\
\bf Family &11.85 &	15.43 & 15.68 & 15.93 & 19.75 & 25.55 & 30.99& 29.13 & 25.56 & 30.12 & 38.02 & 36.42  \\
\hline
\bf Noun-plural &50.23 &	56.68 & 60.56 & 57.96 & 63.21 & 68.92 & 70.35 & 66.52 & 67.87 & 72.97 & 74.02 & 69.14 \\
\bf Jobs &6.73 & 10.52 & 6.82 & 4.04 & 14.39 & 19.78 & 17.68 & 13.30 & 17.59 & 24.24 & 23.06 & 19.36 \\
\bf Verb-past &25.87 &	38.71 & 48.53 & 48.71 & 46.92 & 58.95 & 69.34 & 68.78 & 55.10 & 66.75 & 76.00 & 74.94 \\
\bf Pronouns &5.03 &	6.22 & 7.80 & 7.14 & 10.71 & 12.17 & 15.61 & 15.48 & 13.76 & 16.53 & 19.31 & 19.84 \\
\bf Adj.-gradation &25.00 & 25.00 & 20.00 & 17.50 & 25.00 & 25.00 & 27.50 & 25.00 & 25.00 & 30.00 & 32.50 & 27.50 \\
\bf Nationality 		&0.67 	&1.26 & 0.34 & 0.42 & 2.35 & 2.60 & 1.68 & 2.35 & 3.03 & 3.19 & 3.27 & 2.77 \\
\hline
\end{tabular}
}
\end{table}

\begin{table}[ht!]
\caption{Results for Skip-gram.}\label{tab:skip}
\centering
\resizebox{\textwidth}{!}{
\begin{tabular}{ l||r|r|r|r|r|r|r|r|r|r|r|r }
\hline
\multirow{2}{*}{\bf Type} &\multicolumn{12}{c}{\rule{0pt}{12pt}\bf 3 training epochs}\\
& \bf1\_50 & \bf1\_100 & \bf1\_300 & \bf1\_500 & \bf5\_50 & \bf5\_100 & \bf5\_300 & \bf5\_500 & \bf10\_50 & \bf10\_100 & \bf10\_300 & \bf10\_500\\
\hline
\bf Anton. (nouns) & 0.85& 1.71& 3.34&5.55& 2.20& 3.84& 8.04&10.74& 2.92& 5.41& 9.67&14.08\\
\bf Anton. (adj.) 	&2.26& 3.02& 5.23&8.48 &4.59& 5.69& 9.00&12.37& 6.21& 7.14& 11.32&14.81\\
\bf Anton. (verbs) 	&0.18& 0.36& 0.36& 0.98&0.27& 1.61& 0.45&2.05& 0.89& 1.79& 0.89&2.68\\
\bf State-president &0.18& 0.18& 0.09&0.09 &0.53& 0.71& 0.36&0.62& 0.62& 1.16& 0.71&0.80\\
\bf State-city 		&6.60& 14.26& 8.20&3.48 &17.20& 27.27& 18.89&12.75& 22.99& 33.69& 25.94&21.93\\
\bf Family 			&1.98& 2.72& 2.59&6.79 &3.70& 6.30& 9.01&12.59 &6.30& 8.52& 12.72&16.42\\
\hline
\bf Noun-plural 	&8.11& 14.04& 19.14& 18.77&15.17& 24.62& 27.25&36.41& 18.17& 29.05& 31.23&44.59\\
\bf Jobs 			&1.77& 1.26& 1.09& 1.01&5.05& 3.96& 3.45& 3.53&6.40 &5.81& 4.88&5.39\\
\bf Verb-past 		&1.72& 4.36& 4.14& 6.08&4.20& 8.28& 7.67&12.74 &6.04 &10.62& 9.90&19.97\\
\bf Pronouns 		&0.79& 1.06& 0.66&0.40&2.78& 2.25& 1.72&1.72 &3.97 &4.23& 2.65&2.78\\
\bf Adj.-gradation &2.50& 5.00& 5.00& 10.00&5.00& 7.50& 12.50&17.50 &5.00 &12.50& 12.50&25.00\\
\bf Nationality 		&0.17& 0.08& 0.08&0.00 &0.84& 0.67& 0.17& 0.42&1.26 &1.01& 0.25&0.92\\
\hline
\multirow{2}{*}{} &\multicolumn{12}{c}{\rule{0pt}{12pt}\bf 10 training epochs}\\
& \bf1\_50 & \bf1\_100 & \bf1\_300 & \bf1\_500 & \bf5\_50 & \bf5\_100 & \bf5\_300 & \bf5\_500 & \bf10\_50 & \bf10\_100 & \bf10\_300 & \bf10\_500\\
\hline
\bf Anton. (nouns) & 1.35& 2.63& 6.19&\multicolumn{1}{c|}{x} & 3.27& 5.83& 10.24&\multicolumn{1}{c|}{x} & 4.41& 7.25& 12.23&\multicolumn{1}{c}{x}\\
\bf Anton. (adj.) & 1.74& 4.82& 5.69&\multicolumn{1}{c|}{x}&4.53& 9.12& 10.05&\multicolumn{1}{c|}{x}& 5.57& 11.85& 12.54&\multicolumn{1}{c}{x}\\
\bf Anton. (verbs) & 0.36& 0.00& 0.18&\multicolumn{1}{c|}{x}& 0.98& 1.96& 0.36&\multicolumn{1}{c|}{x} & 1.52& 2.95& 0.62&\multicolumn{1}{c}{x}\\
\bf State-president & 0.27& 0.09& 0.27&\multicolumn{1}{c|}{x}& 1.07& 0.36& 0.80& \multicolumn{1}{c|}{x}&1.52& 0.62& 1.60&\multicolumn{1}{c}{x}\\
\bf State-city & 4.55& 15.15& 9.98&\multicolumn{1}{c|}{x}&14.26& 31.73& 25.85&\multicolumn{1}{c|}{x}&19.88& 39.48& 35.29&\multicolumn{1}{c}{x}\\
\bf Family & 3.09& 3.70& 6.67&\multicolumn{1}{c|}{x}&6.30& 9.14& 13.46&\multicolumn{1}{c|}{x}& 10.37& 12.22& 16.54&\multicolumn{1}{c}{x}\\
\hline
\bf Noun-plural &19.22& 29.95& 23.95&\multicolumn{1}{c|}{x}&31.91& 43.92& 37.91&\multicolumn{1}{c|}{x}&37.39& 47.75& 44.59&\multicolumn{1}{c}{x}\\
\bf Jobs & 2.53& 3.03& 2.53&\multicolumn{1}{c|}{x}&6.99& 7.58& 4.88&\multicolumn{1}{c|}{x} &9.93& 10.44& 7.58&\multicolumn{1}{c}{x}\\
\bf Verb-past & 2.93& 8.25& 8.77&\multicolumn{1}{c|}{x}&7.41& 15.15& 16.69&\multicolumn{1}{c|}{x}&9.73& 18.72& 20.84&\multicolumn{1}{c}{x}\\
\bf Pronouns & 0.66& 0.66& 0.79&\multicolumn{1}{c|}{x}&2.65& 2.25& 3.44&\multicolumn{1}{c|}{x}& 3.84& 3.44& 4.76&\multicolumn{1}{c}{x}\\
\bf Adj.-gradation &2.50& 10.00& 7.50&\multicolumn{1}{c|}{x}&10.00& 15.00& 12.50&\multicolumn{1}{c|}{x}&10.00& 15.00& 15.00&\multicolumn{1}{c}{x}\\
\bf Nationality &0.17& 0.42& 0.08&\multicolumn{1}{c|}{x}&0.50& 1.26& 0.34&\multicolumn{1}{c|}{x}&0.67& 1.60& 0.76&\multicolumn{1}{c}{x}\\
\hline
\end{tabular}
}
\end{table}

\begin{table}[ht!]
\centering
\caption{Results for  GloVe.} \label{tab:glove}
\resizebox{\textwidth}{!}{
\begin{tabular}{ l||r|r|r|r|r|r|r|r|r|r|r|r }
\hline
\multirow{2}{*}{\bf Type} &\multicolumn{12}{c}{\rule{0pt}{12pt}\bf 3 training epochs}\\
& \bf1\_50 & \bf1\_100 & \bf1\_300 & \bf1\_500 & \bf5\_50 & \bf5\_100 & \bf5\_300 & \bf5\_500 & \bf10\_50 & \bf10\_100 & \bf10\_300 & \bf10\_500\\
\hline
\bf Anton. (nouns) &0.36& 1.28& 0.64&0.81& 1.00& 2.92& 1.99&1.72& 1.49& 4.27& 2.63&2.42\\
\bf Anton. (adj.) &0.87& 0.81& 1.34& 1.34&2.44& 4.01& 6.10& 5.81&3.60& 5.40& 8.89&7.62\\
\bf Anton. (verbs) &0.00& 0.00& 0.00& 0.00&0.36& 0.00& 0.00&0.00& 0.36& 0.00& 0.18& 0.00\\
\bf State-president &0.00& 0.00& 0.00&0.00 &0.00& 0.00& 0.00& 0.00&0.00& 0.00& 0.00& 0.00\\
\bf State-city &1.52& 0.98& 1.16&0.98& 3.83& 3.21& 4.01&2.85 &5.17& 4.90& 6.68&5.81\\
\bf Family &3.33& 4.20& 0.99&1.42 &6.67& 6.42& 4.81&3.85 &8.52& 8.64& 7.41&4.35\\
\hline
\bf Noun-plural &14.79& 15.32& 12.69& 5.54&24.47& 26.35& 25.83&14.30 &28.53& 31.46& 33.03&18.70\\
\bf Jobs &0.67& 0.25& 0.00&0.00& 1.43& 0.76& 0.08& 0.00&1.68& 1.09& 0.17&0.00\\
\bf Verb-past &5.39& 6.96& 3.15&0.82 &11.59& 13.71& 7.72&2.78 &15.11& 17.70& 10.80&4.71\\
\bf Pronouns &0.79& 0.66& 0.00&0.00 &1.59& 1.32& 1.46&0.00& 2.12& 1.72& 2.38&0.00\\
\bf Adj.-gradation &7.50& 7.50& 5.00&0.00 &10.00& 12.50& 7.50&7.50& 10.00& 12.50& 10.00&7.50\\
\bf Nationality 		&0.00& 0.00& 0.00& 0.00&0.08& 0.00& 0.00& 0.00&0.08& 0.17& 0.00& 0.00\\
\hline
\multirow{2}{*}{} &\multicolumn{12}{c}{\rule{0pt}{12pt}\bf 25 training epochs}\\
& \bf1\_50 & \bf1\_100 & \bf1\_300 & \bf1\_500 & \bf5\_50 & \bf5\_100 & \bf5\_300 & \bf5\_500 & \bf10\_50 & \bf10\_100 & \bf10\_300 & \bf10\_500\\
\hline
\bf Anton. (nouns) &0.50& 0.85& 1.14& 1.42& 1.28& 2.70& 4.69& 4.05& 1.71& 4.34& 6.33& 5.62\\
\bf Anton. (adj.) &1.68& 2.67& 1.34& 1.34& 3.83& 6.68& 6.56& 6.21& 5.28& 7.96& 9.87& 8.65\\
\bf Anton. (verbs) &0.18& 0.00& 0.00& 0.00& 0.36& 0.18& 0.09& 0.18& 0.89& 0.18& 0.45& 0.36\\
\bf State-president &0.00& 0.00& 0.00& 0.00& 0.00& 0.00& 0.00& 0.00& 0.00& 0.00& 0.00& 0.00\\
\bf State-city &0.98& 1.07& 0.98& 0.45& 3.39& 4.19& 4.01& 2.85& 4.99& 5.97& 7.66& 6.51\\
\bf Family &2.35& 3.70& 2.10& 2.22& 5.43& 5.80& 6.05& 4.20& 7.04& 7.65& 8.52& 5.56\\
\hline
\bf Noun-plural &28.00& 30.56& 15.32& 6.98& 39.79& 43.84& 29.20& 18.02& 43.47& 48.35& 38.44& 28.23\\
\bf Jobs &0.17& 0.00& 0.00& 0.00& 0.59& 0.42& 0.00& 0.00& 0.76& 0.76& 1.18& 0.51\\
\bf Verb-past &7.86& 10.78& 3.98& 1.13& 16.53& 19.25& 10.07& 4.19& 20.82& 23.64& 14.12& 6.81\\
\bf Pronouns &1.32& 1.32& 0.26& 0.00& 3.44& 2.25& 1.06& 0.00& 4.76& 3.57& 1.72& 0.00\\
\bf Adj.-gradation &5.00& 5.00& 5.00& 0.00& 7.50& 10.00& 12.50& 7.50& 15.00& 12.50& 12.50& 7.50\\
\bf Nationality & 0.00& 0.00& 0.00& 0.00& 0.00& 0.00& 0.00& 0.00& 0.00& 0.00& 0.00& 0.00\\
\hline
\end{tabular}
} 
\end{table}

\begin{table}
\centering
\caption{Accuracy on semantic and syntactic part of corpus.} \label{tab:synvssem}
\resizebox{\textwidth}{!}{
\begin{tabular}{ l||r|r|r|r|r|r|r|r|r|r|r|r }
\hline
\multirow{2}{*}{\bf Type} &\multicolumn{12}{c}{\rule{0pt}{12pt} \bf 3  training epochs for CBOW and Skip-gram, 10  training epochs for GloVe.}\\[2pt]
& \bf1\_50 & \bf1\_100 & \bf1\_300 & \bf1\_500 & \bf5\_50 & \bf5\_100 & \bf5\_300 & \bf5\_500 & \bf10\_50 & \bf10\_100 & \bf10\_300 & \bf10\_500\\
\hline
\bf CBOW -- semantics 		&	4.77 & 6.57 & 8.00 & 6.33 & 9.90 & 12.75 & 15.64 & 12.63& 12.11& 15.92 & 19.6& 16.15\\
\bf Skip-gram -- semantics 	&	2.00 & 4.75 & 6.66 &\multicolumn{1}{c|}{x}& 3.71 & 7.57 & 9.62&\multicolumn{1}{c|}{x}& 3.30& 7.62& 10.21&\multicolumn{1}{c}{x} \\
\bf GloVe -- semantics 		&	1.01 & 1.21 & 0.69 &  0.78& 2.38 & 2.76& 2.82& 2.56& 3.19& 3.87& 4.30 &3.63\\
\bf CBOW -- syntactics 		&	11.06 & 15.81 & 19.40 & 16.84 & 17.85 & 22.76 & 26.84& 25.65& 20.48& 26.37  & 30.00& 28.60\\
\bf Skip-gram -- syntactics &	2.51 & 5.51 & 6.81& \multicolumn{1}{c|}{x}& 4.30 & 7.88 & 10.54&\multicolumn{1}{c|}{x}& 5.02& 8.79& 10.24&\multicolumn{1}{c}{x} \\
\bf GloVe -- syntactics 		&	4.86 & 5.11 & 3.50 & 0.98 & 8.20 & 9.11 & 7.10&3.72 & 9.59& 10.77& 9.40 &5.26\\
\hline
\multirow{2}{*}{} &\multicolumn{12}{c}{\rule{0pt}{12pt} \bf 10  training epochs for CBOW and Skip-gram, 25  training epochs for GloVe.}\\[2pt]
& \bf1\_50 & \bf1\_100 & \bf1\_300 & \bf1\_500 & \bf5\_50 & \bf5\_100 & \bf5\_300 & \bf5\_500 & \bf10\_50 & \bf10\_100 & \bf10\_300 & \bf10\_500\\
\hline
\bf CBOW -- semantics 		&	6.77 & 10.90 & 11.42& 10.03 & 13.91 & 19.78 & 22.35 & 19.12& 17.02& 23.23 & 26.90 & 24.20\\
\bf Skip-gram -- semantics 	&	1.89 & 4.40 & 4.83 & 4.23 & 5.07 & 9.69& 10.13& 8.52& 7.21& 12.40& 13.14 & 11.79\\
\bf GloVe -- semantics 		&	0.95& 1.38& 0.93& 0.90& 2.38& 3.26& 3.57& 2.92& 3.32& 4.35& 5.47& 4.45\\
\bf CBOW -- syntax		&	18.92 & 23.07 & 24.01 & 22.63 & 27.10 & 31.24 & 33.69& 31.9& 30.34& 35.61 & 38.03 & 35.59\\
\bf Skip-gram -- syntax &	4.67 & 8.72 & 7.27 & 6.04 & 9.91 & 14.19& 12.63& 12.05& 11.93& 16.16& 15.59& 15.94\\
\bf GloVe -- syntax 		&	7.06& 7.94& 4.09& 1.35& 11.31& 12.63& 8.81& 4.95& 14.14& 14.80& 11.33& 7.17\\
\hline
\end{tabular}
}
\end{table}


How to achieve better accuracy? It was shown in \cite{mikolov2013distributed} that sub-sampling of the frequent words and choosing larger Negative Sampling window helps to improve performance. Also, adding much more text  with information related to particular categories would help (see \cite{pennington2014glove}), especially for class \emph{State-presidents}.

In this paper we had focused more on how number of training epochs influence overall performance in respect to the reasonable time of training and how vector embeddings holds semantics and syntactic information of individual Czech words. We have relatively large corpus for training so we choose 10 iterations (respectively 25 for GloVe) as maximum to compare. To train such models can take more than 3 days with Core i7-3960X, especially for Skip-gram model and vector dimension set to 500. We also do not expect much improvement with more iterations on our corpus, however, we recommend to do more training epochs than it is set by default.

Our goal was not to achieve maximal overall score, but rather to analyze the behavior of word embedding models on Czech language. In following text, we discuss how well these models hold semantic and syntactic information. From results on semantic versus syntactic accuracy (see Table \ref{tab:synvssem}) we can say that for Czech is CBOW approach, which predicts the current word according to the context window better, than predicting a words context based on the word itself as in Skip-gram approach. 

Accuracy on category \emph{State-president} is very low with all models. We would assume to achieve similar results as with category State-city. However, such low score was caused by few simple facts. Firstly, we are missing a data, this is supported by argument that this category has 27\% OOV of questions, than the probability that resulting word will also be missing in vocabulary is going to be high. Second thing is that even if the correct word for a question is not missing in vocabulary, we have more often different corresponding candidates mentioned as presidents of Czech Republic in training data. For example for a question: \emph{"What is a similar word to Czech as is Belarus Alexandr Lukasenko"} we are expecting word \emph{Milos Zeman}, who is our current president. However the models tells us that the most similar word is a word \emph{president}, which is good answer, but we would rather like to see actual name. When we explore other most similar word, we will find \emph{Vaclav Klaus}, who was our former president, fourth similar word was the word \emph{Vaclav Havel}, our first and famous president of Czech Republic after 1992. Based on those statements we can say that we had lack of data corresponding to current presidents in our training corpus.

Czech language has a lot of synonyms for every word that is also why there is overall much better improvement in containing more similar words - TOP 10, rather than just comparing again one word with the highest similarity  TOP 1. Therefore there is a bigger improvement in TOP 1 versus TOP 10 similar words on semantics over it is on syntactic tasks. 

The most interesting results are however for a category Nationality, where we compare nationalities in masculine and feminine form. Complete category is covered in vocabulary. However answers for questions are completely out of topic. For a question which should return feminine form of resident of America, the closest word which model returns is \emph{Oscar Wilde}, respective just his last name, second word is \emph{peacefully philosophy} and another name showing up is \emph{Louise Lasser}. Similar task to category Nationalities with masculine-feminine word form is category Jobs, all models there also perform poorly. This specific task for Czech language seems to be difficult for current state-of-the-art word embeddings methods. 

GloVe model seems to give worse results than Word2Vec models, where on English analogy task gives better accuracy \cite{pennington2014glove}. We would probably get better results with tuning the models properties, but that can be achieved with both presented toolkits.

\section{Conclusion and future work}
 In this paper we introduced new dataset for measuring syntactic and semantic properties of Czech words. We experimented with three state-of-the-art methods of word embeddings, namely, CBOW, Skip-gram, and GloVe. We achieved almost 27\% accuracy on semantic tasks and 38\% on syntactic tasks with our best CBOW model with dimension 300 and exploring top 10 most similar words.

Interesting finding is that on Czech, CBOW model performs much better on word semantics rather than Skip-gram, which performs significantly better on English \cite{mikolov2013distributed}.
 
We made corpus with evaluator script and best trained models publicly available for research purposes at \url{https://github.com/Svobikl/cz_corpus}.

For a future work we would like to further investigate properties of other models for word embeddings and try to use external sources of information (such as part-of-speech tags) during training process.

\section*{Acknowledgements}
This work was supported by the project LO1506 of the Czech Ministry of Education, Youth and Sports and by Grant No. SGS-2016-018 Data and Software Engineering for Advanced Applications.
Computational resources were provided by the CESNET LM2015042 and the CERIT Scientific Cloud LM2015085, provided under the programme "Projects of Large Research, Development, and Innovations Infrastructures.

\bibliographystyle{splncs}
\bibliography{references}

\end{document}